\documentclass[conference]{IEEEtran}
\IEEEoverridecommandlockouts
\usepackage{cite}
\usepackage[table]{xcolor}
\usepackage{amsmath,amssymb,amsfonts}
\usepackage{algorithmic}
\usepackage{graphicx}
\usepackage{textcomp}
\usepackage{xcolor}
\def\BibTeX{{\rm B\kern-.05em{\sc i\kern-.025em b}\kern-.08em
    T\kern-.1667em\lower.7ex\hbox{E}\kern-.125emX}}
\makeatletter
\def\ps@headings{
\let\@oddhead\@empty
\let\@evenhead\@empty
\def\@oddfoot{\@IEEEheaderstyle\hfil\thepage}%
\def\@evenfoot{\@IEEEheaderstyle\thepage\hfil\hbox{}}
}
\def\ps@IEEEtitlepagestyle{
\let\@oddhead\@empty
\let\@evenhead\@empty
\def\@oddfoot{\footnotesize DOI: 10.1109/BigData59044.2023.10386207 ~\copyright 2024 IEEE. Personal use of this material is permitted. Permission from IEEE must be obtained for all other uses.\hfill}%
\let\@evenfoot\@empty
}
\makeatother
\begin{document}

\title{Visual inspection for illicit items in X-ray images using Deep Learning\\
\thanks{The research leading to these results has received funding from the European Union's Horizon Europe research and innovation programme under grant agreement No 101073876 (Ceasefire).}
}

\author{\IEEEauthorblockN{1\textsuperscript{st} Ioannis Mademlis}
\IEEEauthorblockA{\textit{Department of Informatics and Telematics} \\
\textit{Harokopio University of Athens}\\
Athens, Greece \\
imademlis@hua.gr}
\and
\IEEEauthorblockN{2\textsuperscript{nd} Georgios Batsis}
\IEEEauthorblockA{\textit{Department of Informatics and Telematics} \\
\textit{Harokopio University of Athens}\\
Athens, Greece \\
gbatsis@hua.gr}
\and
\hspace{7.2em}\IEEEauthorblockN{3\textsuperscript{rd} Adamantia Anna Rebolledo Chrysochoou}
\IEEEauthorblockA{\hspace{7.2em}\textit{Department of Informatics and Telematics} \\
\hspace{7.2em}\textit{Harokopio University of Athens}\\
\hspace{7.2em}Athens, Greece \\
\hspace{7.2em}adamantia.reb@hua.gr}
\and
\IEEEauthorblockN{4\textsuperscript{th} Georgios Th. Papadopoulos}
\IEEEauthorblockA{\textit{Department of Informatics and Telematics} \\
\textit{Harokopio University of Athens}\\
Athens, Greece \\
g.th.papadopoulos@hua.gr}
}


\maketitle


\begin{abstract}
Automated detection of contraband items in X-ray images can significantly increase public safety, by enhancing the productivity and alleviating the mental load of security officers in airports, subways, customs/post offices, etc. The large volume and high throughput of passengers, mailed parcels, etc., during rush hours practically make it a Big Data problem. Modern computer vision algorithms relying on Deep Neural Networks (DNNs) have proven capable of undertaking this task even under resource-constrained and embedded execution scenarios, e.g., as is the case with fast, single-stage object detectors. However, no comparative experimental assessment of the various relevant DNN components/methods has been performed under a common evaluation protocol, which means that reliable cross-method comparisons are missing. This paper presents exactly such a comparative assessment, utilizing a public relevant dataset and a well-defined methodology for selecting the specific DNN components/modules that are being evaluated. The results indicate the superiority of Transformer detectors, the obsolete nature of auxiliary neural modules that have been developed in the past few years for security applications and the efficiency of the CSP-DarkNet backbone CNN.
\end{abstract}


\section{Introduction}
Detecting contraband items using X-ray scanning of luggage, parcels, etc. is a crucial requirement for ensuring public security (e.g. preventing terrorist attacks, fighting smuggling of illegal goods, etc.) \cite{batsis2023illicit} \cite{Mademlis2024}. X-rays are electromagnetic waves with wavelengths shorter than that of visible light, able to penetrate most materials; X-ray scanners exploit this fundamental property to screen items, such as luggage or packages (e.g., in airports, post/customs offices, etc.). Human operators are able to detect a wide range of potential threats, such as explosives, weapons, or sharp objects, using high-resolution images generated by scanning machines \cite{mery2020x}. However, fully manual screening has important shortcomings: the quality of the scan image can be influenced by several factors, such as occluded objects, cluttered environment or certain material properties of the scanned items \cite{akcay2022towards}, while heavy traffic during rush hours may mentally overload human security officers. Thus, illicit items may be missed, due to the need for ``the line to keep moving" or because of perceptual limitations. The high volume and high throughput of X-ray scans in such scenarios render manual screening ineffective and demand automated Big Data analysis solutions.

Efficient automated X-ray image analysis/screening for illicit item detection is nowadays possible thanks to the advances of computer vision and machine learning. Such methods are ideal for large-scale information processing and, therefore, hold the potential to facilitate the detection of illicit item trafficking activities, suspected terrorist attacks, etc. Deep Neural Networks (DNNs) have proven to be remarkably capable in supporting human operators for similar tasks, thus greatly increasing their productivity and reducing the possibility of mistakes. Both whole-image recognition and object detection methods have been proposed for illicit/contraband item detection in X-ray images. While the former ones simply classify an entire image and assign it an overall class label, algorithms of the latter type identify Regions-of-Interest (RoIs), i.e., bounding boxes that localize (in 2D pixel coordinates) specific objects visible in an input image. While there have been significant advancements in object detection algorithms through the use of DNNs, achieving sufficient performance in real-world scenarios continues to be a challenge \cite{Thermos_2017_CVPR} \cite{Gkountakos2019} \cite{Liu2020} \cite{mademlis2023vision}.

The typical goal is for a deployed DNN to automatically detect illicit goods, such as drugs or weapons, in passengers, luggage or mailed parcels. The dominant trends are similar to those of the RGB image analysis, but obviously different training datasets are utilized. Additionally, special/auxiliary neural modules are commonly employed as part of the overall DNN architecture, so that accuracy is improved in the face of typically encountered issues such as high occlusions, very cluttered backgrounds and large class imbalance. These mechanisms are designed to handle similar application domain-specific aspects.

Given the practical importance of the task, recent literature surveys have overviewed detection of illicit items in X-ray scans for security applications \cite{rafiei2023computer} \cite{wu2023object}. Yet, none of them has assessed the various state-of-the-art methods using a common experimental evaluation protocol, thus rendering cross-method performance comparisons difficult. In an attempt to remedy the situation, this paper contributes a thorough quantitative assessment, using the most common relevant public dataset (SIXray \cite{miao2019sixray}) and comparing various combinations of DNN backbones, auxiliary modules and detection heads. The results indicate:
\begin{itemize}
\item the superiority of Transformer detectors,
\item the obsolete nature of auxiliary neural modules that have been developed in the past few years for security applications, 
\item the high efficiency of the CSP-DarkNet backbone CNN.
\end{itemize}

The remainder of the paper is organized as follows. Section \ref{sec:RelatedWork}  overviews the recent literature on illicit item detection in X-ray scan images using DNNs. Section \ref{sec:Methods} briefly presents the specific DNN backbones, auxiliary modules and detection heads that are being quantitatively assessed. Section \ref{sec:eval} outlines the experimental evaluation process, which was conducted on a well-known public dataset, and discusses the obtained results. Section \ref{sec:concl} concludes the preceding discussion by identifying the implications of these findings and directions for future research.

\section{Related Work}
\label{sec:RelatedWork}
Various approaches have been employed over the years for illicit items detection in X-ray scan images. The method of \cite{akccay2016transfer} addressed the issue of limited training data by employing a pretrained CNN and fine-tuning it in the X-ray domain. This is an important issue in automated X-ray screening, since negative images (where no illicit item is present) are typically significantly more than the positive ones, with this fact reflected in the relevant available datasets.

Common DNNs for object detection have also been evaluated with regard to their discrimination capacity and transferability between different X-ray scanners \cite{gaus2019evaluating}; examples include Faster R-CNN \cite{ren2015faster}, Mask R-CNN \cite{he2017mask} and RetinaNet \cite{lin2017focal}. However, modifying fast, anchor-based, single-stage object detectors such as Single Shot MultiBox Detector (SSD) \cite{liu2016ssd} or You Only Look Once (YOLO) \cite{redmon2016you} is the most common approach, due to their ability to operate in real-time even in embedded computer hardware. Such modifications may have various forms. For instance, a Cascaded Structure Tensor (CST) is proposed in \cite{hassan2020cascaded} which takes advantage of contour-based information to extract object proposals; the latter ones are then classified using a CNN. An alternative lightweight object detector, called LightRay, is introduced in \cite{ren2022lightray} as a modified version of the YOLOv4 model for small illicit item detection in complex backgrounds. It consists of a fast MobileNetV3 \cite{howard2019searching} backbone CNN and a feature enhancement network that includes a Lightweight Feature Pyramid Network (LFPN) \cite{lin2017feature}, to obtain information of objects at different scales, and a Convolutional Block Attention Module (CBAM) \cite{woo2018cbam}, for refining feature maps through a spatial attention mechanism.

A different approach is followed in \cite{shao2022exploiting}, where a novel mechanism called Foreground and Background Separation (FBS) is proposed for separating illicit items from complex/cluttered backgrounds. This is achieved by using a feature extraction DNN combined with Spatial Pyramid Pooling (SPP) and a Path Aggregation Network, which extracts high-level features. These feature maps serve as an input to two neural decoders, which reconstruct the background and the foreground simultaneously. Then, an attention module directs the overall model's focus on the foreground objects.

Focusing on real-time performance, YOLOv5 is modified in \cite{song2022improved} using the Stem \cite{wang2018pelee} and CGhost \cite{han2020ghostnet} modules, resulting in a model with reduced number of parameters that still achieves competitive results in comparison with the baseline method. 

\section{Employed Methods for Illicit Item Detection in X-Ray Scan Images}
\label{sec:Methods}
This section briefly illustrates the different deep neural modules/architectures that have been selected for comparative experimental assessment. First, the relevant one-stage object detection heads are described. Then, the various backbone networks and the auxiliary modules that have been included are being presented. Finally, the specific combinations of the above-mentioned components that are experimentally compared in Section \ref{sec:eval} are discussed and justified.

\subsection{Detection Heads}
Most single-stage object detectors utilize reference anchor boxes of different sizes and aspect ratios, which are placed at various positions across the input image. The goal of these anchor boxes is to capture the variation in object shapes and sizes present in the dataset. Typically, they are predefined (e.g., calculated based on prior knowledge of the sizes, aspect ratios, and distributions of ground-truth objects in the COCO dataset \cite{lin2014microsoft}). In many implementations the match between these predefined anchor boxes and the training dataset is verified before training commences, by computing the achievable recall rate if the object detector using these anchors has access to the ground-truth for all objects in the dataset. If this recall rate is too low, the predefined anchors are assumed to be unfit and a new set of dataset-specific anchor boxes is estimated (e.g., via clustering). The detection head essentially outputs the offset (in pixel space) of each predicted bounding box from a known anchor box. After a set of raw detections has been generated, a typically handcrafted Non-Maximum Suppression (NMS) algorithm refines them by merging/filtering any spatially overlapping detected RoIs which correspond to a single visible object \cite{batsis2023illicit} \cite{shepley2023confluence} \cite{Symeonidis2023}.

\begin{figure}[htbp]
\centerline{\includegraphics[width=\linewidth]{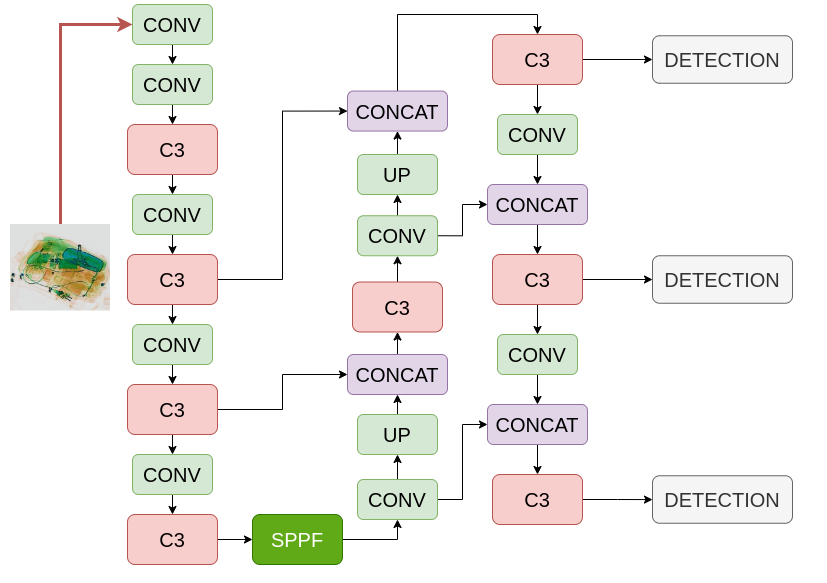}}
\caption{YOLOv5 overall architecture.}
\label{yoloarch}
\end{figure}

You Only Look Once (YOLO) \cite{redmon2016you} is a series of fast anchor-based, single-stage object detectors, where object localization and classification are performed using a single CNN. This architecture can, however, be divided into a backbone network, a succeeding neck network and a final prediction head. YOLOv5 \cite{jocher2020yolov5}, which is an update of YOLOv4 \cite{bochkovskiy2020yolov4}, is inspired by EfficientNet \cite{tan2019efficientnet} and, thus, can be easily reconfigured for different network complexity profiles. Out of the common variants (YOLOv5s, YOLOv5m, YOLOv5l, YOLOv5x) the one employed in this paper is YOLOv5l. The overall YOLOv5 architecture is presented in Fig. \ref{yoloarch}.

\begin{figure*}
\centerline{\includegraphics[width=\linewidth]{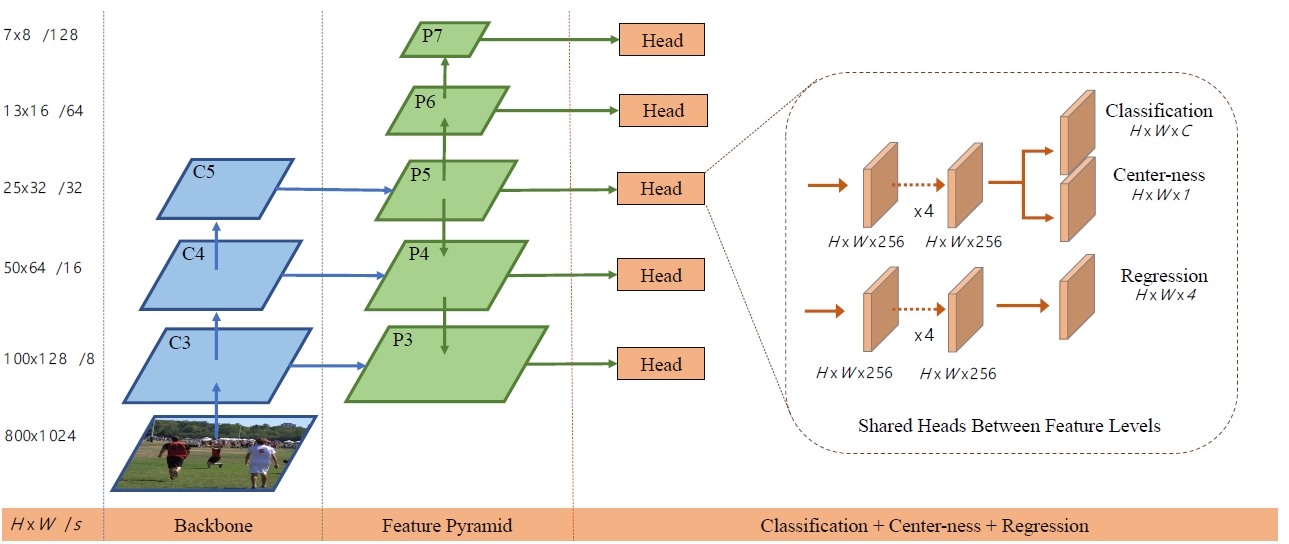}}
\caption{The architecture of FCOS \cite{tian2019fcos}.}
\label{fig:fcos}
\end{figure*}

While all detectors of the YOLO family rely on preset anchors, Fully Convolutional One-stage Object Detection (FCOS) \cite{tian2019fcos} is one of the first successful anchor-free one-stage CNN detectors that outputs per-pixel predictions. Thus, it avoids the initial computational load for setting-up the anchors before the main training process, as well as all relevant hyperparameters that are difficult to tune. FCOS requires a neck network based on Feature Pyramid Network (FPN) \cite{lin2017feature}, which aggregates different backbone-derived feature maps corresponding to different image scales. Features from the downsampling path are fed to the upsampling one through lateral synapses. Thus, objects of different sizes can be detected at different levels of the feature pyramid. Detection is conducted by the shared head, which analyzes the outputs of the FPN levels and is composed of three branches: one for classification, one for centerness and one for regression. All three of them output per-pixel predictions: the first one predicts the object's class, the second one how far a pixel deviates from the center of its associated bounding box, while the third one outputs the dinstance (in pixels) of the pixel in question and the corners of its bounding box. One disadvantage of FCOS is that it requires higher input image resolutions to operate correctly, due to the per-pixel nature of its predictions; this creates an execution time overhead during both the training and the inference stage. A high-level diagram of its architecture is depicted in Fig. \ref{fig:fcos}.

The anchor-free direction is also followed by YOLOv8 \cite{yolov8}, a recent successor to YOLOv5 that directly predicts the centers of bounding boxes. Along with various minor improvements in the CNN architecture and an enhanced data augmentation strategy during training, YOLOv8 achieves an outstanding balance between inference speed and prediction accuracy. The YOLOv8 variant that is utilized in this paper is YOLOv8l.

\begin{figure*}[htbp]
\centerline{\includegraphics[width=\linewidth]{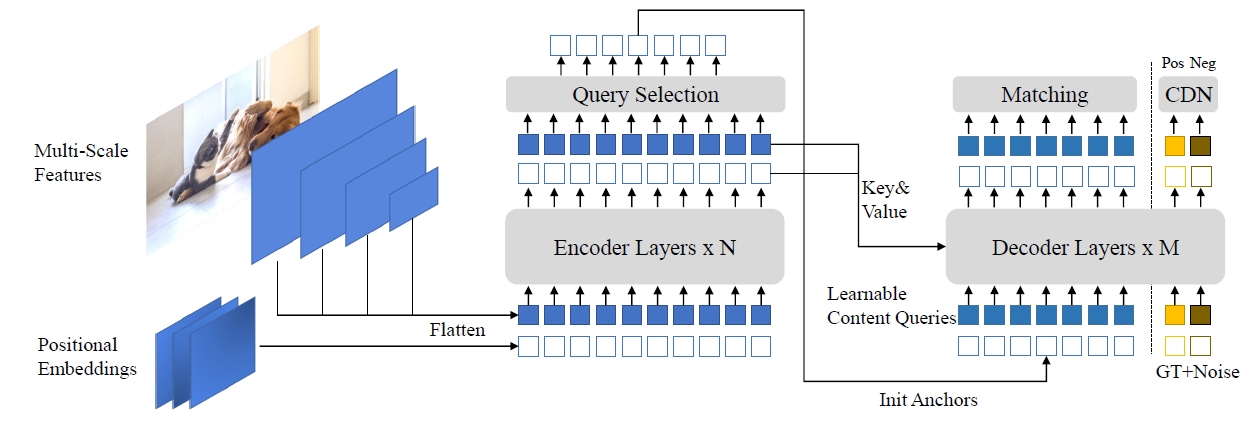}}
\caption{The architecture of DINO \cite{zhang2022dino}.}
\label{fig:dino}
\end{figure*}

Despite the early dominance of CNNs as detection heads, top-performing Vision Transformer DNNs have emerged during the past few years. One of the first such approaches was Detection Transformer (DETR) \cite{carion2020detr}: it is an Encoder-Decoder Transformer DNN \cite{ilia}, placed after a CNN backbone, which treats image blocks as tokens. DETR handles object detection as a set prediction task and assigns labels by bipartite graph matching. Learned positional encodings, the so-called ``object queries", essentially look for a particular object in the image. The method is not only anchor-free, but also NMS-free; DETR does not need any handcrafted algorithmic components. A state-of-the-art improvement of DETR is DINO \cite{zhang2022dino}, which accumulates various minor enhancements over baseline DETR and reinstates the use of anchor boxes in a Transformer-compatible manner. Moreover, it exploits an additional contrastive loss term during training \cite{hadsell2006}, by adding two different types of noise to the same ground-truth RoI; the resulting bounding box with a smaller/larger amount of noise is considered a positive/negative sample, respectively. The goal is to push the DNN towards avoiding duplicate bounding box outputs that correspond to a single ground-truth object. A high-level diagram of the DINO architecture is depicted in Fig. \ref{fig:dino}.

\subsection{Backbone Networks}
ResNet-101 \cite{he2016deep} is a well-known CNN backbone, very commonly employed for almost any image analysis task. It is one of the first CNNs that was able to be trained with large network depth without being negatively impacted by the gradient vanishing problem, mainly thanks to its introduction of the ``skip synapses". The continuing popularity of ResNet for almost a decade showcases its value for the wider computer vision community.

\begin{figure*}[htbp]
\centering
\centerline{\includegraphics[width=0.7\linewidth]{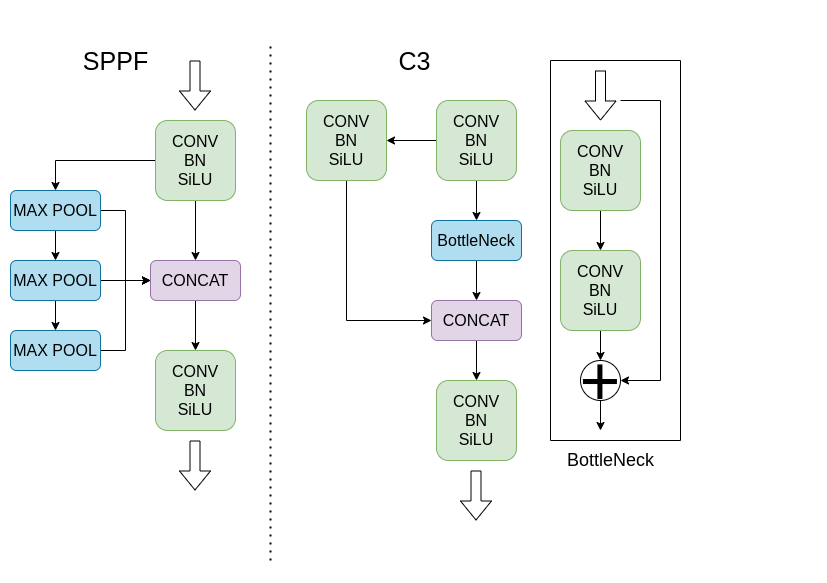}}
\caption{The main CSP-Darknet53 components.}
\label{yolocomp}
\end{figure*}

The default backbone CNN of YOLOv5 is CSP-Darknet53 \cite{bochkovskiy2020yolov4}, a modified version of Darknet53 \cite{redmon2018yolov3} combined with a Cross Stage Partial Network (CSPNet) strategy \cite{wang2020cspnet}, which is specifically designed for assisting object detection. As presented in Fig. \ref{yolocomp}, the main convolutional block of CSP-Darknet53 consists of convolutional layers, residuals and the SiLU activation function, while the final feature maps are refined using a Spatial Pyramid Pooling-Fast (SPPF) module \cite{he2015spatial}. The neck network consists of a Feature Pyramid Network (FPN) and a Path Aggregation Network (PAN) \cite{liu2018path}. These modules repeatedly fuse feature maps from different scales and depth levels, thus leading to final image representations, which are simultaneously characterized by accurate spatial localization details, rich semantics and high invariance regarding object detection. Finally, the prediction head outputs the candidate detected RoIs through a set of convolutional operations.

Due to the rather low inference speed of very deep ResNet variants, fast generic CNN backbones appeared over the years, targeting execution on embedded computers with limited processing power. One of the most important relevant architectures is MobileNet, which accelerates inference by incorporating ``separable convolutions" \cite{howard2017mobilenets}. Additionally, the widespread use of $1 \times 1$ convolutional kernels allows their optimized implementation through generalized matrix multiplication, while test accuracy and training are aided by the utilization of batch normalization and ReLU activation functions. MobileNetV2 improves this architecture by periodically decimating the number of convolutional channels along the depth dimension (similarly to SqueezeNet \cite{iandola2016squeezenet}), adding skip synapses and reducing the need for greater number of channels per convolutional layer in later layers. The next iteration, i.e., MobileNetV3 \cite{howard2019searching}, further enhances the architecture by introducing a channel-wise attention module within each separable convolution and optimizing architectural details at the network design phase, through the use of Neural Architecture Search (NAS). Overall, MobileNets achieve a very good balance between speed and accuracy: in most applications, they lag only slightly compared to non-lightweight deep CNN backbones, while being significantly faster.

An alternative lightweight fast CNN backbone is EfficientNet \cite{tan2019efficientnet}. As in the case of MobileNetV3, it is designed by employing NAS based on reinforcement learning; however, the reward function prefers a low total number of computational operations during the forward pass instead of a low required inference runtime. In general, however, the individual neural layers/modules are similar to the ones utilized by MobileNetV3. EfficientNet variants of various complexities are available, so that the speed-accuracy trade-off can be adjusted based on the desired application and the computational power which is available at the inference stage. More complex variants are typically deeper (more convolutional layers), wider (more channels per layer) and process input images of higher resolution. The architecture family has been improved with EfficientNetV2 \cite{tan2021efficientnetv2}, which makes NAS to also reward higher training efficiency and incorporates enhancements in the regularization scheme utilized during training. The EfficientNet variant utilized in this paper is EfficientNetV2-S.

\subsection{Auxiliary Modules}
Due to the peculiarities of illicit item detection in X-ray scan images of luggage, parcels, etc., various additional domain-specific, plug-in neural modules have been proposed over the years. For instance, the method of \cite{miao2019sixray} introduces a module called Class-balanced Hierarchical Refinement (CHR), to enhance the prediction capacity of the CNN under extreme class imbalance. CHR can be placed as a neck module on top of any CNN backbone.

In an orthogonal direction, the De-occlusion Attention Module (DOAM) \cite{wei2020occluded} is a neural module designed to overcome occlusions in X-ray images; this is important because occlusions are common, due to the absorption of X-rays by certain materials, such as metals, and the visual overlap of multiple objects within densely packed parcels. The latter phenomenon implies that a single pixel may correspond to multiple semantic classes, of objects located at different vertical distances from the sensor, due to the penetrative nature of X-rays. Thus, the overall X-ray image can be considered a superposition of various sub-images. DOAM consists of two sub-modules, named Edge Guidance (EG) and Material Awareness (MA), which identify edge and material cues for all visible objects. An alternative domain-specific module is Lateral Inhibition Module (LIM) \cite{tao2021towards}, which includes two components: Bidirectional Propagation (BP) and Boundary Activation (BA). The former one minimizes the impact of neighboring regions, by isolating irrelevant information and the latter one captures object boundaries. Both DOAM and LIM have shown promising results in overcoming object occlusion issues in X-ray scan images.

In a subsequent attempt to overcome the issues induced by the typically high visual overlap of objects within a densely packed luggage/parcel, the method in \cite{Ma2022} introduces the so-called Dense De-overlap Module (DDoM). It operates by assigning learned weights to each channel of a convolutional feature tensor, indicating how relevant it is to the object class in question. This operates under the assumption that different convolutional channels are responses to different sub-images, including irrelevant background ones. Finally, the integrated Prohibited Object Detection (POD) method \cite{ma2023occluded} for X-ray image analysis combines a learnable Gabor layer for edge information retrieval, a spatial attention module for directing focus on low-level features, a Global Context Feature Extraction (GCFE) module and a Dual Scale Feature Aggregation (DSFA) module to enhance semantic information from high-level features.

\subsection{Methodology for comparative assessment}
\label{ssec::comparative}
The literature of DNNs for illicit item detection in X-ray scan images mostly employs common neural architectures/building blocks (detectors, backbones, necks), typically preferring fast and proven ones. Thus, most of the specific neural components reviewed in the previous subsections were chosen to be included in this comparative experimental assessment because they are commonly found in recent relevant papers (e.g., YOLOv5, FCOS, ResNet-101, EfficientNet, MobileNet). However, the final selection of individual components is influenced by other considerations as well, such as state-of-the-art status (e.g., YOLOv8, DINO). In particular, one the goals of this work is to identify how relevant domain-specific neural modules, such as CHR, LIM, DOAM or DDoM, remain in the face of the advancements offered by modern generic detectors.

Thus, given that it would be very impractical to quantitatively evaluate all potential combinations of the selected neural building blocks, the following process has been followed:
\begin{itemize}
    \item First, commonly employed CNN detectors are evaluated in combination with the selected CNN backbones.
    \item Second, the state-of-the-art one-stage CNN detector, i.e., YOLOv8, is evaluated in combination with the best-performing CNN backbone.
    \item Third, the various auxiliary modules (serving as neck subnetworks) are evaluated in combination with the overall best CNN detection head.
    \item Four, the best performing CNN backbone is evaluated in combination with DINO; a representative of state-of-the-art Transformer-based detection heads.
\end{itemize}

The details and the results of this incremental experimental assessment are presented in Section \ref{sec:eval}.

\section{Experimental Evaluation}
\label{sec:eval}
This section overviews the common experimental setup used for evaluating and comparing the components presented in Section \ref{sec:Methods}. Subsequently, the assessment results are reviewed and discussed.

\subsection{Experimental Dataset}
\begin{figure}[htbp]
\centerline{\includegraphics[width=1\linewidth]{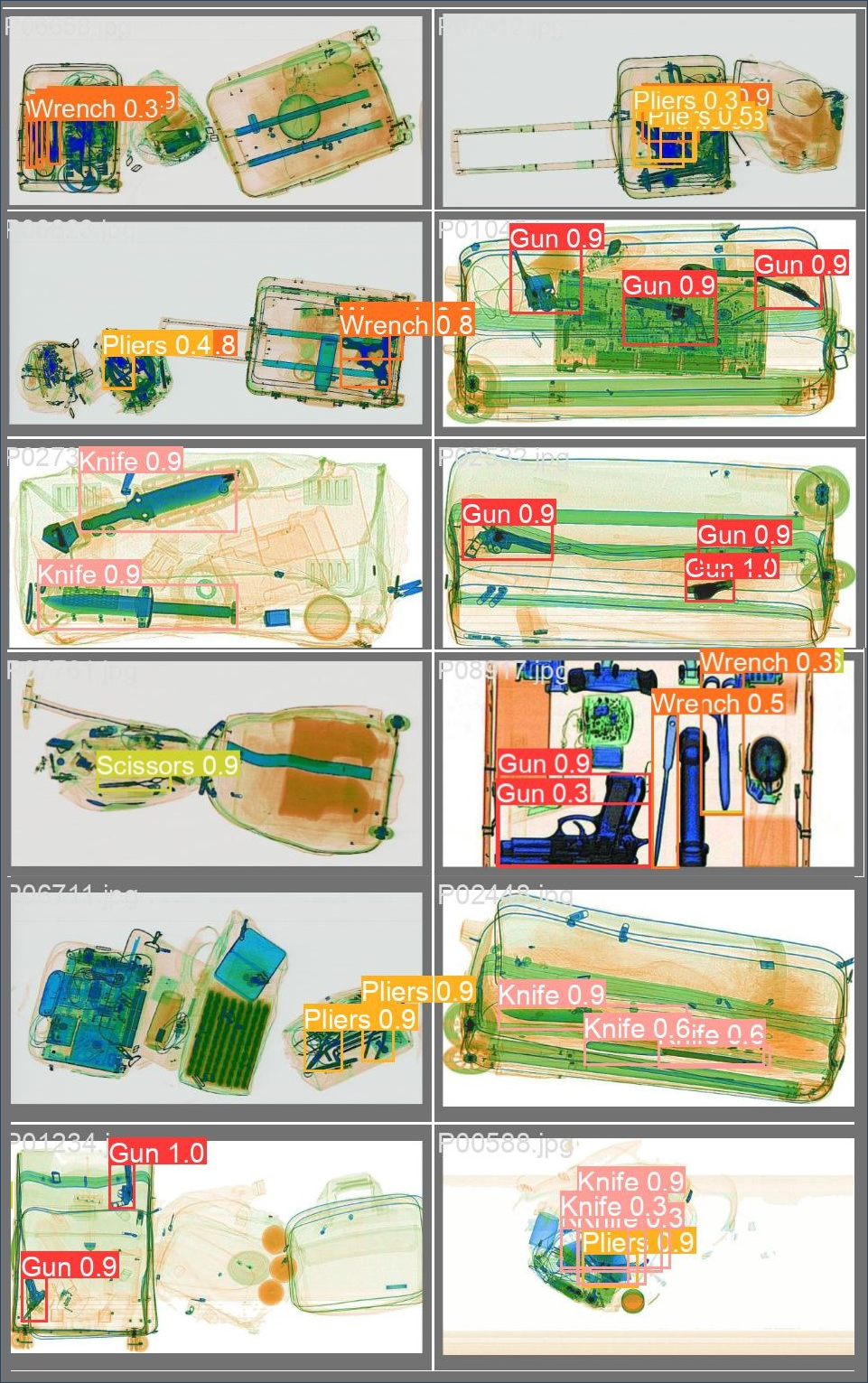}}
\caption{Predictions on the SIXray test subset.}
\label{fig::preds}
\end{figure}

SIXray \cite{miao2019sixray} is employed for conducting the experimental method assessment. It is a well-known publicly available X-ray security dataset consisting of 1,059,231 X-ray images from subway stations. The 6 classes of illicit objects contained in these images are ``gun", ``knife", ``wrench", ``pliers" and ``scissors". Additionally, a ``negative" class includes all images without any illicit item. Three different dataset subsets are typically utilized in different experimental setups, namely SIXray10, SIXray100 and SIXray1000, where the number indicates the ratio of negative against positive samples. SIXray contains ground-truth whole-image class label annotations manually set by human security inspectors, while their ground-truth object RoIs/bounding boxes are available only for the test set. This paper uses the revised object detection annotations for the training subset provided by \cite{nguyen2022towards}. Despite the fact that only images containing at least one contraband item were utilized, the official training-test set split was adopted. Fig. \ref{fig::preds} depicts examples of detections on SIXray test set images.

\subsection{Evaluation Metrics}
The effectiveness of the proposed method is measured using the mean Average Precision (mAP) metric. In object detection tasks, IoU is commonly used to measure the overlap between the predicted and the corresponding ground-truth RoI. In addition, a threshold value is defined in order to decide whether the prediction is actually correct. True Positives (TP), False Positives (FP), and False Negatives (FN) depend on the IoU, the predicted label and the ground-truth label. These elementary metrics are utilized to calculate Precision and Recall:

\begin{equation}
Precision = \frac{TP}{TP+FP}\label{prec}.
\end{equation}

\begin{equation}
Recall = \frac{TP}{TP+FN}\label{rec}.
\end{equation}

\noindent The Precision-Recall (PR) curve depicts the trade-off between precision and recall for different discrimination thresholds. Average Precision (AP) is the area under the PR curve and its range is between 0 to 1. AP is defined as:

\begin{equation}
AP =  \int_{0}^{1} p(r) \,dr\label{AP}.
\end{equation}

\noindent mAP is calculated as the mean of AP over all classes:

\begin{equation}
    mAP = \frac{1}{N} \sum_{i}^{N}AP_i\label{mAP}.
\end{equation}

\subsection{Experimental Evaluation}
Evaluation of all competing method combinations in the SIXray dataset was conducted using the mAP metric at a 0.5 IoU threshold.

\begin{table*}
\centering
\begin{tabular}{|l|l|l|}
\hline \cellcolor{green!25}
\textbf{Detector}&\cellcolor{green!25} \textbf{Backbone Architecture} &\cellcolor{green!25} \textbf{mAP} \\ \hline
\textit{YOLOv5}       & CSPDarkNet-53                                & 0.82              \\ \hline
                      & ResNet-101                                    & 0.81 \\ \hline
                      & MobileNetV3                                  & 0.76  \\ \hline
                      & EfficientNetV2-S                             & 0.81 \\ \hline
\textit{FCOS}         & ResNet-101                                    & 0.78  \\ \hline
                      & MobileNetV3                                  & 0.73  \\ \hline
                      & EfficientNetV2-S                             & 0.73  \\ \hline
\textit{YOLOv8}       & CSPDarkNet-53                                & 0.84   \\ \hline

\textit{DINO}         & CSPDarkNet-53                               & \textbf{0.89} \\ \hline\cellcolor{blue!25}&\cellcolor{blue!25}&\cellcolor{blue!25}\\\hline
                     \cellcolor{green!25}\textbf{Detector}&\cellcolor{green!25} \textbf{Auxiliary Module}     & \cellcolor{green!25} \textbf{mAP}\\ \hline
\textit{YOLOv8}       & CHR & 0.68  \\ \hline
                      & LIM & 0.82 \\ \hline
                      & DOAM & 0.78 \\ \hline
                      & DDoM        & 0.81  \\ \hline

\end{tabular}
\label{tab:Res}
\caption{Results of the quantitative assessment of the various selected method combinations, under the chosen experimental protocol. mAP@0.5 is the employed evaluation metric (higher is better).}
\end{table*}

Table I summarizes the mAP of the evaluated method combinations, selected under the rationale described in Subsection \ref{ssec::comparative}. As it can be seen, the Transformer-based DINO outperforms all CNN-based detectors, but the CSPDarkNet-53 CNN backbone, which has been designed specifically for object detection, surpasses all competing approaches. Finally, as it can be deduced from the quantitative results, the domain-specific auxiliary modules that have been evaluated as neck subnetworks in combination with YOLOv8/CSPDarkNet-53 are essentially useless in combination with such an advanced CNN detector; they significantly degrade its accuracy. One potential reason may be that they are not really generic plug-in modules able to augment any CNN backbone/detector combination, but can only cooperate effectively with specific such combinations. Exploring this aspect is a fertile future research avenue.

It must be noted that the above results contradict those of the survey in \cite{wu2023object}, which concludes that Transformer-based DNNs do not work equally well on X-ray images because they emphasize contours, while CNNs emphasize texture. In practice, the comparative assessment results presented in this paper indicate that this is not in fact an issue, at least when a CNN backbone is utilized in combination with a state-of-the-art Transformer detector. This is in-line with the recent findings of \cite{Isaac2023}, where Transformer-based detection heads are shown to outperform all competitors.

\section{Conclusions}
\label{sec:concl}
The automated detection of contraband items in X-ray images obtained in airports, subways or post/customs offices is a task critical for public safety. Due to the large volume and high throughput of passengers, mailed parcels, etc., this is a Big Data analysis problem that requires fast algorithms. Existing one-stage DNNs for object detection have indeed been adapted and trained for this application domain, but so far they have not been compared under a common evaluation protocol. This paper presented exactly such a comparative assessment of various commonly employed or state-of-the-art deep neural components for object detection (detection heads, backbones, auxiliary domain-specific necks), using a well-known, large-scale public relevant dataset. The results indicate the superiority of Transformer detectors, the obsolete nature of auxiliary neural modules that have been developed in the past few years for security applications and the high efficiency of the CSP-DarkNet backbone CNN. Future research directions include an investigation of whether domain-specific auxiliary modules can be effectively utilized in combination with advanced modern object detectors to further improve accuracy, as well as how an end-to-end Transformer solution would perform in comparison to the winning CSP-DarkNet+DINO combination.

\section*{Acknowledgment}
The research leading to these results has received funding from the European Union's Horizon Europe research and innovation programme under grant agreement No 101073876 (Ceasefire). This publication reflects only the authors views. The European Union is not liable for any use that may be made of the information contained therein.

\bibliographystyle{IEEEtran}
\bibliography{bibliography.bib}

\end{document}